\documentclass[iicol]{sn-jnl}

\usepackage{biblatex}
\usepackage{orcidlink}
\usepackage{colortbl}%
\usepackage{array}%
\usepackage{bm}%
\usepackage[export]{adjustbox}%
\usepackage[column=0]{cellspace}%
\usepackage{graphicx}%
\usepackage{multirow}%
\usepackage{amsmath,amssymb,amsfonts}%
\usepackage{amsthm}%
\usepackage{mathrsfs}%
\usepackage[title]{appendix}%
\usepackage[dvipsnames]{xcolor}%
\usepackage{textcomp}%
\usepackage{manyfoot}%
\usepackage{booktabs}%
\usepackage{algorithm}%
\usepackage{algorithmicx}%
\usepackage{algpseudocode}%
\usepackage{listings}%
\begin{document}

\title[Article Title]{Localized Control in Diffusion Models via Latent Vector Prediction}
%%Advanced Image Generation through Localized Conditioning in Diffusion Models
%%Local Control in Image Generation via Adapting Conditional Diffusion Models

%%=============================================================%%
%% GivenName	-> \fnm{Joergen W.}
%% Particle	-> \spfx{van der} -> surname prefix
%% FamilyName	-> \sur{Ploeg}
%% Suffix	-> \sfx{IV}
%% \author*[1,2]{\fnm{Joergen W.} \spfx{van der} \sur{Ploeg} 
%%  \sfx{IV}}\email{iauthor@gmail.com}
%%=============================================================%%

\author*[1,2]{\fnm{Pablo} \sur{Domingo-Gregorio} \orcidlink{0009-0000-3506-8898}} \email{pablo.domingo.gregorio@upc.edu}

\author[1]{\fnm{Javier} \sur{Ruiz-Hidalgo} \orcidlink{0000-0001-6774-685X}}\email{j.ruiz@upc.edu}
%%TODO\equalcont{These authors contributed equally to this work.}

\affil[1]{\orgdiv{Image Group}, \orgname{Universitat Politècnica de Catalunya}, \orgaddress{\city{Barcelona}, \country{Spain}}}

\affil[2]{\orgdiv{R\&D}, \orgname{Napptilus Tech Labs}, \orgaddress{\city{Barcelona}, \country{Spain}}}

%%TODO\affil[2]{\orgdiv{Department}, \orgname{Organization}, \orgaddress{\street{Street}, \city{City}, \postcode{10587}, \state{State}, \country{Country}}}

%%==================================%%
%% Sample for unstructured abstract %%
%%==================================%%

\abstract{Diffusion models emerged as a leading approach in text-to-image generation, producing high-quality images from textual descriptions. However, attempting to achieve detailed control to get a desired image solely through text remains a laborious trial-and-error endeavor. Recent methods have introduced image-level controls alongside with text prompts, using prior images to extract conditional information such as edges, segmentation and depth maps. While effective, these methods apply conditions uniformly across the entire image, limiting localized control. In this paper, we propose a novel methodology to enable precise local control over user-defined regions of an image, while leaving to the diffusion model the task of autonomously generating the remaining areas according to the original prompt. Our approach introduces a new training framework that incorporates feature-level masking and a penalization loss term, which leverages the prediction of the initial latent vector at any diffusion step to enhance the correspondence between the current step and the final sample in the latent space. Extensive experiments demonstrate that our method effectively synthesizes high-quality images with controlled local conditions. The code will be available once the paper is published.
}


\maketitle

\section{Introduction}\label{intro}
In recent years, large-scale Diffusion models \cite{dalle, imagen, latentdiff, scalablediff, rectifiedflow} have become the new paradigm for text-to-image tasks. However, textual descriptions often prove to be insufficient in conveying the desired level of detail in the generated image. Following the saying ``a picture is worth a thousand words'', recent methods, such as Text-to-Image Adapter (T2I-Adapter) and ControlNet \cite{t2iadapter, controlnet} have been proposed to incorporate image-level spatial conditions such as edges, segmentations, and depth maps into text-to-image generation. These methods are designed to apply conditions across the entire image, obeying the entirety of the prior condition map. Although these advances have gained significant attention for enabling more controllable image synthesis, they typically rely on conditioning inputs that span the entire image. As a result, fine-grained, localized control remains challenging, since the model requires prior information not only for the region of interest but also for the background. 

\begin{figure*}
    \centering
    \begin{tabular}{0{wc{0.18\textwidth}} | 0{wc{0.18\textwidth}} | 0{wc{0.18\textwidth}} | 0{wc{0.18\textwidth}}}
         \toprule \toprule
		\noalign{\smallskip}
            \textbf{Prompt} & \multicolumn{3}{c}{\textbf{\textit{`A living room with a sofa.'}}}\\
            \midrule
            \textbf{Models} & \cellcolor[rgb]{0.918,0.663,0.290}\textbf{T2I-Adapter} & \cellcolor[rgb]{0.451,0.518,0.812}\textbf{ControlNet} & \cellcolor[rgb]{0.451,0.565,0.526}\textbf{Ours} \\
            \midrule
            \textbf{Original Image} & \multicolumn{3}{c}{\textbf{Generated Images}} \\
            \includegraphics[width=2.65cm, valign=c]{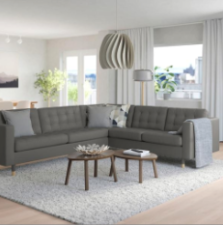}&
            \includegraphics[width=2.65cm, valign=c]{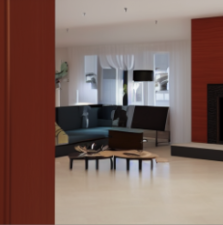}&
            \includegraphics[width=2.65cm, valign=c]{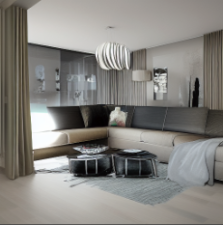}&
            \includegraphics[width=2.65cm, valign=c]{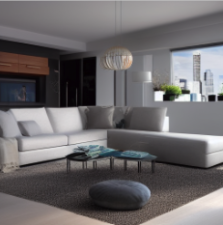}\\
            \midrule
            \textbf{Input Edges} & \multicolumn{3}{c}{\textbf{Generated Edges}} \\
            \includegraphics[width=2.65cm, valign=c]{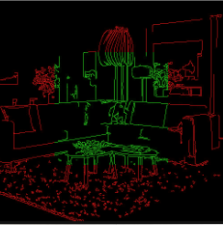}&
            \includegraphics[width=2.65cm, valign=c]{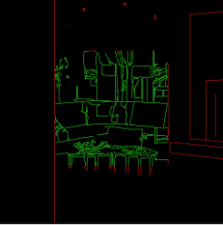}&
            \includegraphics[width=2.65cm, valign=c]{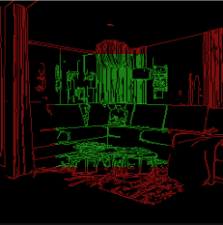}&
            \includegraphics[width=2.65cm, valign=c]{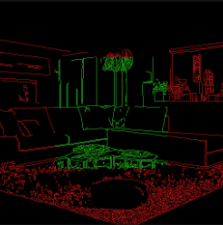}\\
            \midrule \bottomrule
    \end{tabular}
    \caption{Comparison of generated images and their edges from current image-level models given the prompt “A living room with a sofa.” The first column shows the original image and its edges. The subsequent columns present the generated images and their corresponding edge maps. Edge images are divided into (\textcolor{ForestGreen}{green areas}), which represent edges within the region of interest where a spatial condition is applied, and we aim to preserve edge similarity; and (\textcolor{red}{red areas}), which denote unconditioned regions where the model is expected to generate a coherent background with natural contours, without over or under generation.}
    \label{fig:comparison}
\end{figure*}

Conventional approaches to controlling the entire image often fall short of the fine-grained image manipulation demanded by users and industries. In many cases, only partial information about an image is available, yet we aim to complete it as a coherent whole. For instance, in design applications, one may want to vary the environment around a static object or character without altering its structure. This limitation can be observed in Figure \ref{fig:comparison}, where some models produce incomplete images when given partial conditioning or tend to generate content beyond the intended conditioned region.

For example, T2I-Adapter \cite{t2iadapter} interprets the absence of edges as smooth or contourless areas, focusing exclusively on the region of interest while neglecting the rest of the image. This produces backgrounds that lack significant detail, often leaving them almost empty.
In contrast, ControlNet \cite{controlnet} generates more complete images, but struggles to adhere to the specified conditions within the region of interest. Instead of maintaining the desired structure, ControlNet frequently introduces additional edges within this area. This deviation highlights its limitations in localizing the conditions precisely, despite its ability to enrich the overall image context. These behaviours are further analyzed through qualitative and quantitative comparisons in section \ref{sec:results} to support these observations.

This raises the question: \textit{\textbf{Can we develop a method that controls structural information within a region of interest while still generating complete images?}} In this paper, we explore this scenario and introduce a methodology to locally control user-defined regions while ensuring the model generates cohesive and complete images. Our approach operates directly in the latent space of the diffusion model, where, building on prior work, we guide the generation by predicting the initial latent vector at a given time step to satisfy image-level conditions through a loss function. This enables localized control without requiring conditioning over the full image, offering users the flexibility to anchor specific regions while still producing complete, high-quality images.

Our contributions can be summarized as follows. 
\begin{itemize} 
    \item We introduce a novel approach for local control within diffusion generation, enabling fine-grained manipulation with image-level conditions. 
    \item We propose a masking-based strategy to localize spatial information, balancing control within a defined region while maintaining creative freedom outside it. 
    \item We enforce the similarity between the initial latent vector and its prediction by introducing a novel penalization loss. This loss ensures precision in masked regions while generating detailed and coherent backgrounds. 
\end{itemize}

\section{Related Work}\label{relwork}
The field of text-to-image synthesis has progressed along two primary paths: one aimed at improving image quality and generation speed, and the other at increasing controllability through conditioning. While both have led to significant advances, this section focuses on the latter, exploring conditioning techniques that enable more precise and localized control.

Diffusion models are inherently controllable through textual prompts. Although the alignment between text and image continues to improve as models advance, there remains a significant gap in controlling fine-grained details during generation. To address this limitation and facilitate more controllable image generation, recent works have focused on three main areas: the addition of customized concepts \cite{taming, dreambooth, instantbooth}, the ability to generate images containing multiple concepts \cite{multiconcept, fastcomposer, collage}, and conditional control over the generation \cite{t2iadapter, ipadapter, gligen, controlnet, controlnet++, smart_control, text_driven_regions, ctrl-x}. 

In the case of achieving customized concepts, for instance, Dreambooth \cite{dreambooth} explored the personalization of new concepts into the model's knowledge through fine-tuning the Diffusion model, using a small sub-set of user-provided image examples of that concept. Other methods \cite{taming, instantbooth} have focused their efforts mainly on reducing the number of examples needed and avoiding fine-tuning by introducing new concepts through encoders instead. While these approaches are effective for obtaining the desired content of the generated image, they do not grant the user direct control over the final composition and appearance of the image.

Another challenge in textual prompting arises when multiple subjects are included in the prompt. Often, one subject gains disproportionate attention, leading the model to neglect the other. To address this issue, methods like Multiple Concept Control \cite{multiconcept} have been using cross-attention layers, to ensure a balanced representation of all subjects in the generated image. Other approaches, \cite{fastcomposer, collage, gligen} improve control over the composition by linking specific subject embeddings to image-level positions using masks or bounding boxes into the attention maps. Although greater control can be achieved with the latter models, they often lack the ability to define fine-grained details within those areas.

\label{sota:spatial_control}In order to achieve more fine-grained control over the text-to-image synthesis, methods like T2I-Adapter and ControlNet \cite{t2iadapter, controlnet} propose manipulating the latent feature space by introducing image-level features through the training of specialized encoders. This approach enables for precise control over the structure and composition of generated images but lacks the ability to locally control them. 

\begin{figure*}
    \centering
    \includegraphics[width=13.5cm]{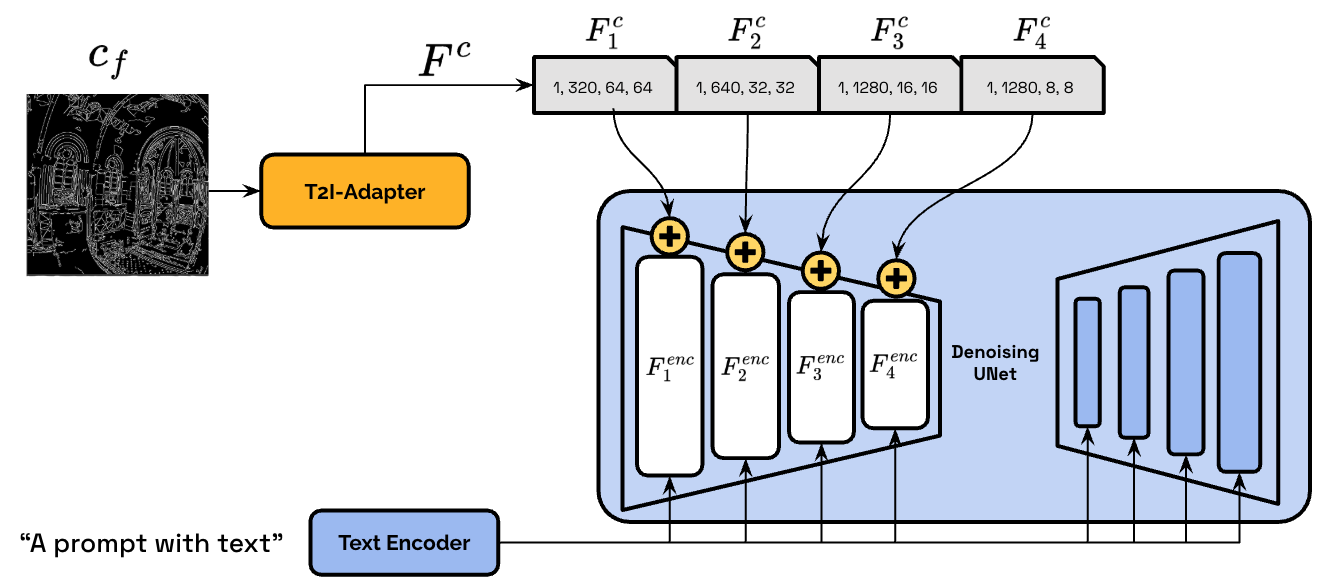}
    \caption{Overview of the T2I-Adapter integration. The input condition $c_f$ is processed by the adapter network $F_{\text{AD}}(c_f)$ to generate condition features $F^c = \left\{F^c_1, F^c_2, F^c_3, F^c_4\right\}$, which are designed to match the spatial dimensions of the corresponding encoder feature maps $F^{\text{enc}} =  \left\{F^{\text{enc}}_1, F^{\text{enc}}_2, F^{\text{enc}}_3, F^{\text{enc}}_4\right\}$ produced by the UNet. At each encoder layer, the adapter features are fused with the UNet features via element-wise addition, enabling the integration of image-level conditioning information into the diffusion process.}
    \label{fig:adapter_functioning}
\end{figure*}

Some methods have extended T2I-Adapter \cite{t2iadapter} and ControlNet \cite{controlnet} to improve structural or appearance alignment, but without enabling explicit localization control. ControlNet++ \cite{controlnet++} introduces a reward model that evaluates how well the generated image matches the input condition, encouraging global consistency but lacking the ability to focus control on a specific region. CTRL-X \cite{ctrl-x} combines structure and appearance guidance by integrating findings from Image-Prompt Adapter (IP-Adapter) \cite{ipadapter}, yet both the reference and structural images influence the entire output, limiting flexibility in background generation. SmartControl \cite{smart_control} proposes modulating ControlNet’s skip connections to resolve conflicts between condition and prompt, learning a spatially-varying control strength. However, control still applies across the whole image and cannot be explicitly localized. Lastly, Text-Driven Image Editing \cite{text_driven_regions} identifies salient objects to anchor control, but localization is determined automatically rather than by the user, and the method is specialized for object manipulation, making background freedom infeasible.

Unlike prior approaches, which apply conditions globally or rely on model-driven localization, our method enables explicit, user-defined fine-grained manipulation of specific image regions. This is achieved by operating in the latent space and integrating image-level conditions through a dedicated loss function, allowing for precise control within targeted areas while preserving coherent and high-quality outputs elsewhere. Importantly, our design remains compatible with existing models and leverages the efficiency of T2I-Adapter, making it both practical and computationally accessible for deployment across various settings.

\begin{figure*}
    \centering
    \includegraphics[width=16.5cm]{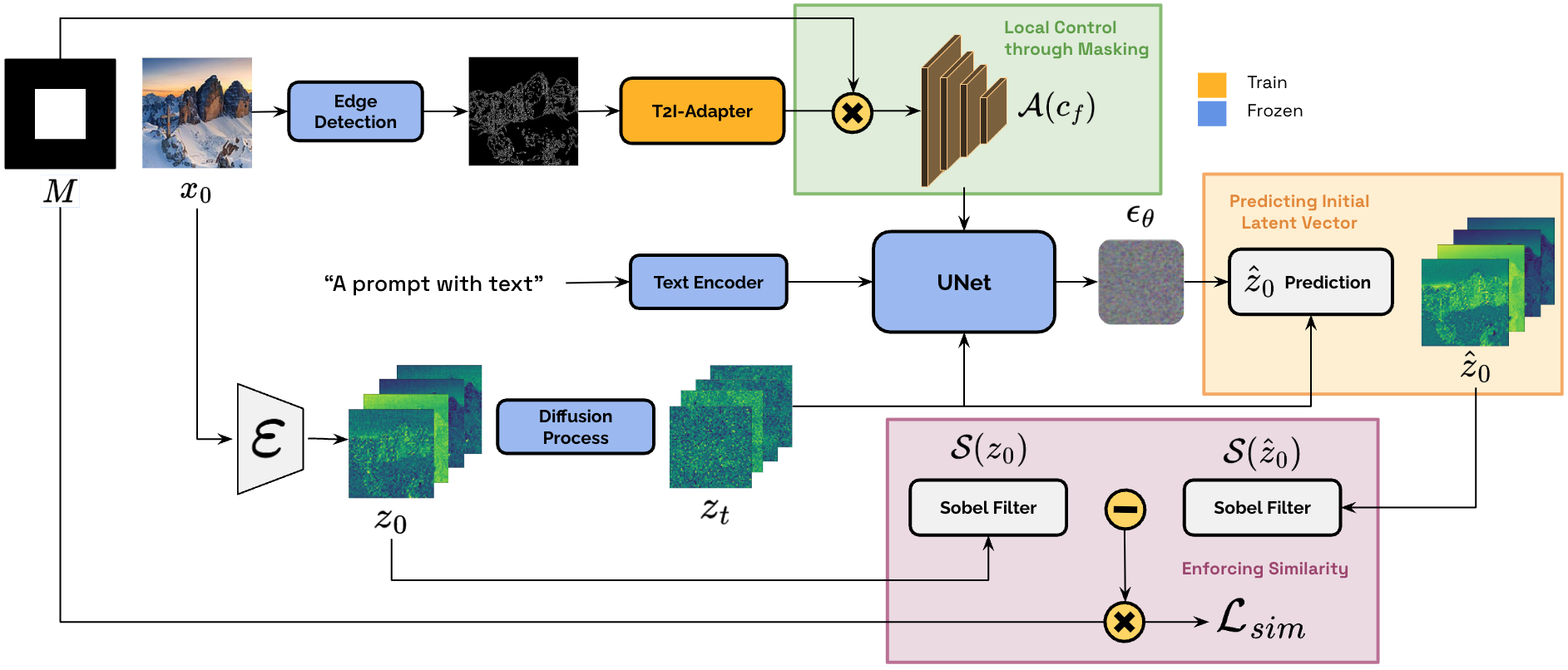}
    \caption{Overview of our training methodology at each step $t$ of the denoising process. Given the textual prompt $p$, the control condition $c_{f}$ and region of interest defined by the mask $M$, an image \textit{I} is passed through $\mathcal{E}$ to obtain its latent representation $z_{0}$. At each training step, we apply \textbf{Local Control} (see Section \ref{sec:local_control}) to the T2I-Adapter features. A timestep $t$ is then selected, and diffusion is performed on $z_{0}$ to generate the noisy input $z_{t}$. Latent variable $z_{t}$ jointly with $y$ and the T2I-Adapter features are passed into the denoising network $U$, to predict $\epsilon_{\theta}$. We subsequently \textbf{Predict the initial latent vector} $\widehat{z}_{0}$ (see Section \ref{sec:predicting_z0}) using $\epsilon_{\theta}$ and $z_{t}$ to \textbf{Enforce Similiarity} (see Section \ref{sec:enforcing_similarity}) by applying a Sobel filter $\mathcal{S}$ on the original and predicted latents $z_{0}$, $\widehat{z}_{0}$ within the region of interest.}
    \label{fig:overview}
\end{figure*}

\section{Methodology}\label{method}
\subsection{Preliminaries}
In this work, we build upon Stable Diffusion \cite{latentdiff}, a variant of diffusion models that operates in the latent space using pre-trained variational autoencoders (VAE) comprising an encoder $\mathcal{E}$ and decoder $\mathcal{D}$. The encoder $\mathcal{E}$ maps an input image \textit{I} into a latent representation, while the decoder $\mathcal{D}$ reconstructs images from latent vectors. This latent-space formulation offers a significant efficiency advantage over pixel-based diffusion models like Imagen \cite{imagen} due to its lower dimensionality.

The generative process in Stable Diffusion consists of two key stages: diffusion and denoising. In the diffusion process, an input image \textit{I} is encoded into its initial latent representation $z_{0}$ using the encoder $\mathcal{E}$ as follows:
\begin{equation}
    z_{0} = \mathcal{E}(\textit{I})
\end{equation}
The initial latent representation $z_{0}$ is then progressively noised using a Gaussian noise $\epsilon \sim \mathcal{N}(0, \boldsymbol{\mathcal{I}})$ over a series of steps \textit{T}, resulting in a sequence of noisy latents $z_{t}$ at each timestep $t$ and culminating in indistinguishable Gaussian noise $z_{T}$. The denoising process then aims to reverse this transformation by predicting the added noise $\epsilon$ at each step, through a learnable model $U$ conditioned on the noisy latent $z_{t}$ at timestep $t$ and an optional text prompt $c_{p}$ derived from the Contrastive Language-Image Pre-training (CLIP) \cite{clip} text encoder.
\begin{equation}
    \epsilon_{\theta} = U(z_{t}, t, c_{p})
\end{equation}
During training, the model $U$ learns to predict the added noise at each step by minimizing the difference between the actual noise $\epsilon$ and the predicted noise $\epsilon_{\theta}$ expressed as the following training objective \eqref{eq:diff_loss}.
\begin{equation}
    \mathcal{L}_{\text{diff}} = \mathbb{E}_{z_{0}, t, c_{p}, \epsilon} ||\epsilon - \epsilon_{\theta}||^{2}
    \label{eq:diff_loss}
\end{equation}

Once trained, $U$ can generate image latents by starting with Gaussian noise and iteratively refining it through denoising, guided by text prompts $c_{p}$. Sampling efficiency can be improved using fast sampling techniques such as \cite{ddim, dpm_solver, unipc}. 

This work focuses on T2I-Adapter models \cite{t2iadapter}, which provide a modular and lightweight solution by integrating image-level features into the latent space representation, as shown in Figure \ref{fig:adapter_functioning}. Specifically, T2I-Adapter extracts condition features $F^c = \{F^c_1, F^c_2, F^c_3, F^c_4\}$ from the input condition $c_f$ using a shallow network $F_{\text{AD}}(c_f)$. These features are designed to match the shape and resolution of the encoder feature maps $F^{\text{enc}} = \{F^{\text{enc}}_1, F^{\text{enc}}_2, F^{\text{enc}}_3, F^{\text{enc}}_4\}$ produced by the UNet. At each encoder layer $i$, the adapter features are fused with the UNet features via simple element-wise addition.

\begin{equation}
    \widehat{F}^{\text{enc}}_i = F^{\text{enc}}_i + F^c_i
    \label{eq:t2i-adapter_features}
\end{equation}

The integration of adapter features at each encoder layer enables the model to inject external spatial cues into the generation process. During training, the objective in equation \eqref{eq:new_objective} is modified to incorporate this conditioning.

\begin{equation}
    \mathcal{L}_{\text{diff}} = \mathbb{E}_{z_{0}, t, c_{p}, c_{f}, \epsilon} ||\epsilon - \epsilon_{\theta}||^{2}
    \label{eq:new_objective}
\end{equation}

\begin{figure*}
    \centering
    \includegraphics[width=14.5cm]{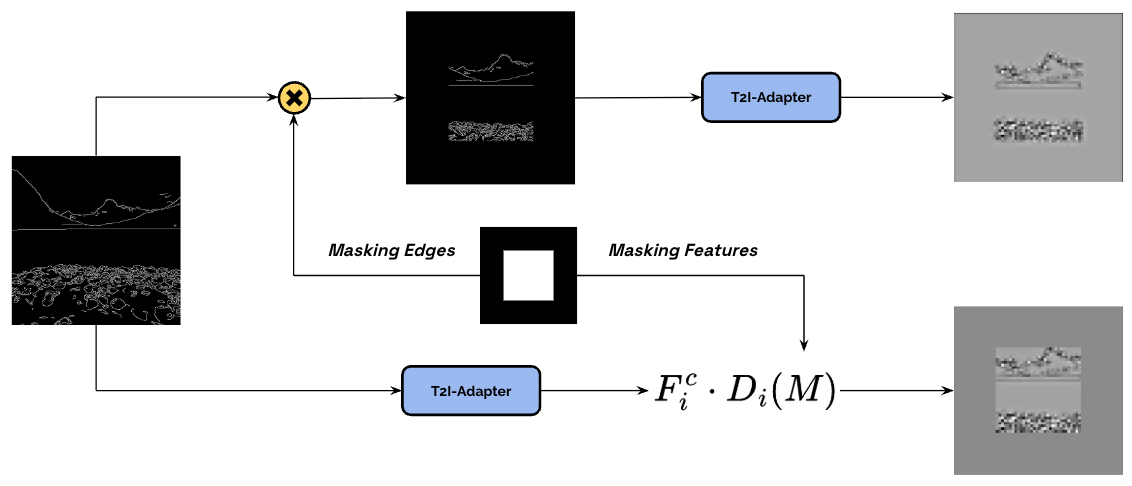}
    \caption{Comparison between direct masking of the edge map and our feature-level masking approach. When the mask is applied directly to the edge map, the model cannot distinguish between unstructured regions (intended smooth areas) and unconditioned regions (where no guidance should be applied), resulting in a loss of control outside the masked area. In contrast, our approach applies the mask to the T2I-Adapter features, preserving three distinct zones: structured regions with edge information, unstructured regions intended to be smooth, and unconditioned regions outside the mask. This separation enables precise, localized control while allowing the model to generate coherent and diverse content beyond the masked area.}
    \label{fig:feature_distinction}
\end{figure*}

\subsection{Overview}
We propose a method for integrating localized spatial conditioning into the diffusion process, enabling precise, user-defined control over specific regions of the image. Our goal is to guide the generation using localized edge-based structural cues, while allowing the model to freely generate content elsewhere. To achieve this, we modify both the conditioning mechanism and the training objective, ensuring that localized control can be enforced without compromising image quality or coherence.

Our methodology, illustrated in Figure \ref{fig:overview}, consists of three key components: 1) \textbf{Local Control through Masking} (see Section \ref{sec:local_control}), which ensures that spatial information is applied within the mask while allowing creative freedom outside of it; 2) \textbf{Predicting Initial Latent Vector} at each training step (see Section \ref{sec:predicting_z0}), which provides a reference for the intended output; and 3) \textbf{Enforcing Similarity} (see Section \ref{sec:enforcing_similarity}), which uses the $\widehat{z}_{0}$ prediction to enforce precision and consistency within the masked regions. By doing so, we ensure that the model adheres to localized spatial conditions while generating detailed, coherent images that align with the overall prompt.

\subsection{Local Control through Masking}\label{sec:local_control}
We aim to locally guide the structure of generated images using edge maps as spatial conditions. While prior methods like T2I-Adapter inject global structural information effectively, achieving localized control by constraining only specific regions, remains challenging.

In this work, we define the \textit{region of interest} (ROI) as the specific spatial area of the image where the local structural guidance is applied. The mask is used to specify the boundaries of the ROI, but is not itself the ROI, rather it serves to select the relevant area for conditioning.

\begin{figure*}
    \centering
    \begin{tabular}{0{wc{0.18\textwidth}} | 0{wc{0.18\textwidth}} | 0{wc{0.18\textwidth}} | 0{wc{0.18\textwidth}} | 0{wc{0.18\textwidth}}}
         \toprule \toprule
		\noalign{\smallskip} 
             \textbf{Masked} &
             \multicolumn{2}{c|}{\textbf{Masking Edges}} & \multicolumn{2}{c}{\textbf{Masking Features}} \\
             \textbf{Condition} & \cellcolor[rgb]{0.918,0.663,0.290}\textbf{T2I-Adapter} & \cellcolor[rgb]{0.451,0.565,0.526}\textbf{Ours} & \cellcolor[rgb]{0.918,0.663,0.290}\textbf{T2I-Adapter} & \cellcolor[rgb]{0.451,0.565,0.526}\textbf{Ours} \\
             \midrule
             \includegraphics[width=2.75cm, valign=c]{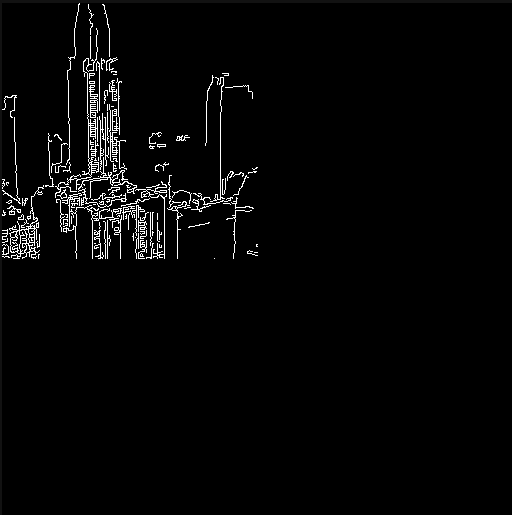} &
             \includegraphics[width=2.75cm, valign=c]{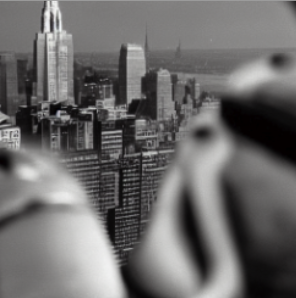} &
             \includegraphics[width=2.75cm, valign=c]{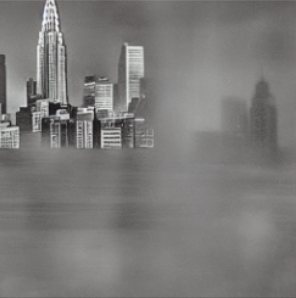} &
             \includegraphics[width=2.75cm, valign=c]{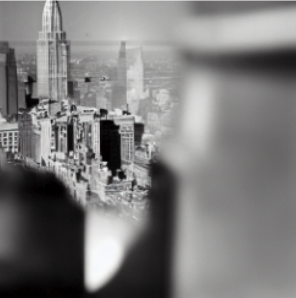} &
             \includegraphics[width=2.75cm, valign=c]{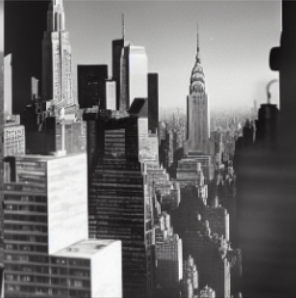} \\
             \includegraphics[width=2.75cm, valign=c]{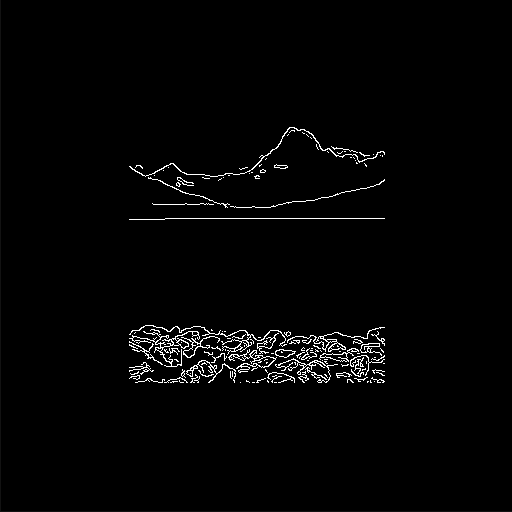} &
             \includegraphics[width=2.75cm, valign=c]{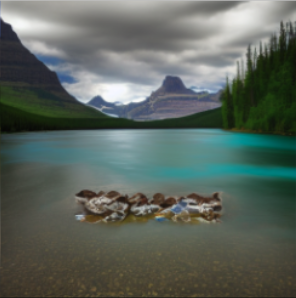} &
             \includegraphics[width=2.75cm, valign=c]{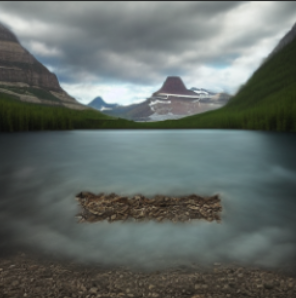} &
             \includegraphics[width=2.75cm, valign=c]{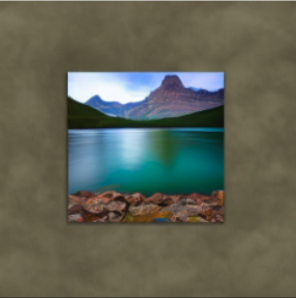} &
             \includegraphics[width=2.75cm, valign=c]{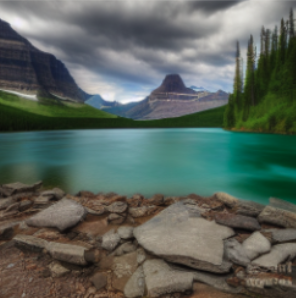} \\
             \midrule \bottomrule
    \end{tabular}
    \caption{Comparison of the effect of applying the mask either to the edge condition (Masking Edges) or to the adapter features (Masking Features) for both T2I-Adapter and our method. When the mask is applied to the edge maps, both models interpret areas without contours as smooth, resulting in blurred or empty regions outside the masked area. In contrast, when the mask is applied to the adapter features, T2I-Adapter continues to treat unconditioned areas as contourless, while our method correctly distinguishes between the ROI and the background. This allows our approach to preserve structure inside the mask and generate more coherent, realistic backgrounds beyond it.}
    \label{fig:masking_difference}
\end{figure*}

To enable genuine localized control, the conditioning input must encode three distinct zones: (1) structured regions within the ROI, where edges define the desired local contours; (2) unstructured regions within the ROI, where the absence of edges indicates smooth or contourless areas; and (3) unconditional regions outside the ROI, where the model has creative freedom. This can be achieved by applying the mask not to the input edge map directly, but to the features produced by the T2I-Adapter, as illustrated in Figure \ref{fig:feature_distinction}, and retraining the model to learn this separation of control zones. In doing so, the T2I-Adapter learns to encode partial and spatially localized control that the U-Net can correctly interpret during the denoising process. This approach preserves the distinction between regions with intended structure, regions intended to be smooth, and areas where no conditioning is provided.

In contrast, masking the ROI in the edge map, preserving structural cues only within the mask and blacking out the rest, leads to a critical issue. Since T2I-Adapter is trained on full edge maps, it cannot differentiate between “no structure present” and “no condition present”, interpreting masked or blacked-out regions as contourless or smooth. As a result, the model faithfully follows the structure in the masked-in region but renders the rest of the image as smooth or empty. As shown in Figure \ref{fig:masking_difference}, applying direct masks to the edge maps causes the model to interpret blacked-out areas as smooth, whereas feature-level masking enables our approach to properly distinguish between partially conditioned and unconstrained regions. This results in more natural image synthesis compared to models trained with T2I-Adapter on full edge maps.

The integration of these masked features follows the same mechanism as in the original T2I-Adapter. Formally, the features from the adapter are first masked, using a downsampled version of the binary mask $M$ to match the spatial resolution of each layer, and then injected into the UNet feature maps $F^{\text{enc}}$ at each layer $i$. The updated feature maps are computed as \eqref{eq:adapted_features}. This ensures that conditioning is only applied in the intended regions, while leaving the rest of the image unconstrained by structural guidance.
\begin{equation}
    \widehat{F}^{\text{enc}}_i = F^{\text{enc}}_i + (F^c_i \cdot D_i(M))
    \label{eq:adapted_features}
\end{equation}
During training, the spatial condition $c_{f}$ is obtained by applying the Canny edge detection algorithm \cite{canny_detector} to the image, and the mask is created by selecting a randomized region of at least 10,000 pixels for each sample.

\subsection{Predicting Initial Latent Vector}\label{sec:predicting_z0}
While training diffusion models, the denoising process centers on predicting the noise added at each step, rather than reconstructing the image directly. As a result, the model’s output during training is noise, which complicates enforcing penalties or constraints during training, especially when attempting to guide localized features such as edges.

To overcome this, we leverage the formulation from \cite{hodenoising}, which shows that it is possible to go from $z_{t}$ to a version of $z_{0}$ which we call predicted initial latent vector $\widehat{z}_{0}$, at any diffusion timestep during training. By predicting $\widehat{z}_{0}$, we can obtain a reference point that allows us to re-enter the feature space after each noise prediction during backpropagation, enabling us to apply penalties that enforce desired spatial conditions in specific regions. This method shifts the focus from noise to feature comparison, providing a clearer basis for guiding desired properties and enforcing penalties in the image generation process.

Specifically, from the diffusion process,
\begin{equation}
    q(z_{t}|z_{0}) = \mathcal{N}(z_{t};\sqrt{\overline{\alpha}_{t}}\cdot z_{0}, (1-\overline{\alpha}_{t})\cdot \mathcal{I})
    \label{eq:ho_eq}
\end{equation}
Where $\overline{\alpha}_{t}$ defines the amount of noise added at each step $t$, we can use the standard Gaussian reparameterization $\mathcal{N}(\mu,\sigma^{2}) \rightarrow z = \mu + \sigma \cdot \mathcal{E},  \mathcal{E} \sim \mathcal{N}(0, \boldsymbol{\mathcal{I}})$ to obtain an expression of $z_t$
\begin{equation}
    z_{t} = \sqrt{\overline{\alpha}_{t}} \cdot z_{0} + \sqrt{1-\overline{\alpha}_{t}} \cdot \mathcal{E}
    \label{eq:zt}
\end{equation}
Isolating $z_{0}$ would in principle allow us to recover $z_0$ at any training step. However, during training, $z_{t}$ is a noisy version of $z_{0}$, and exact reconstruction is generally not achieved due to the inherent stochasticity of the diffusion process and small prediction errors. Therefore, the value obtained is an estimate, denoted as $\widehat{z}_0$. 
\begin{equation}
    \widehat{z}_{0} = \frac{z_{t} - \sqrt{1-\overline{\alpha}_{t}}\cdot \mathcal{E}}{\sqrt{\overline{\alpha}_{t}}}
    \label{eq:pred_equation}
\end{equation}
Even though $\widehat{z}_0$ is an estimate, it provides a suitable basis for applying penalties and enables more precise spatial control during training.

\subsection{Enforcing Similarity}\label{sec:enforcing_similarity}
Building on the $\widehat{z}_{0}$ prediction derived in the previous section, we aim to enforce edge alignment within the ROI, thereby improving the T2I-Adapter's localized structural guidance. However, since $z_{0}$ is a latent representation of the generated image rather than its structure, it is necessary to extract edge features in the latent space before applying any penalty. To achieve this, we apply a high-pass filter to both the predicted $\widehat{z}_{0}$ and the original $z_{0}$ and compare them to apply the corresponding penalty.

We selected the Sobel filter to extract structural (edge) information, as Sobel \cite{sobel} is differentiable and thus suitable for gradient-based optimization, unlike non-differentiable alternatives such as Canny. This choice is further supported by an ablation study comparing Sobel to alternative filters, such as Laplace, which confirmed Sobel’s effectiveness for our purposes. The Sobel filter is defined in \eqref{eq:sobel}, where $W$ is a $3 \times 3$ kernel and $\bigodot$ denotes convolution.
\begin{equation}
    \mathcal{S}(z) = W_{3\times3} \bigodot z 
    \label{eq:sobel}
\end{equation}
During training this filter is applied to both $\widehat{z}_{0}$ and $z_{0}$ to obtain their edge representations in latent space. These filtered features are then masked by $M$ (see Section \ref{sec:local_control}) to focus the comparison on the ROI. The Mean Square Error (MSE) between the masked Sobel-filtered features of $\widehat{z}_{0}$ and $z_{0}$ is then computed as a loss term to penalize discrepancies in edge structure. This penalty encourages the Adapter to align structural content more precisely within the target area.

Mathematically, given the Sobel-filtered outputs $\mathcal{S}(z_{0})$ and $\mathcal{S}(\widehat{z}_{0})$, the penalization loss within the masked area is calculated as follows, where $\mathit{M}$ denotes the ROI.
\begin{equation}
     \mathcal{L}_{\text{sim}} = M \cdot || \mathcal{S}(z_{0}) - \mathcal{S}(\widehat{z}_{0})||^2
\end{equation}
To integrate the penalization into the global objective while training, the overall loss function is defined as the sum of the standard diffusion loss $\mathcal{L}_{\text{cond}}$ and the edge similarity loss prediction-based $\mathcal{L}_{\text{sim}}$, combined with a regularization term $\lambda$ to balance the penalization contributions. By doing so, the model is trained to enforce spatial information accurately within the ROI.
\begin{equation}
    \mathcal{L}_{\text{total}} = \mathcal{L}_{\text{diff}} + \lambda \cdot \mathcal{L}_{\text{sim}}
\end{equation}
\section{Experiments}\label{experiments}
\subsection{Dataset}
To train the T2I-Adapter we used the 600k subset of the LAION-Aesthetics V2 \cite{laion} public datasets, comprised of 600K pairs of text-image samples. For our purposes, we pre-processed the dataset to remove any pairs where the image resolution was lower than 256 pixels in any dimension. The remaining pairs were passed through the canny edge algorithm to get the edge content. For validation and test purposes, we reserved 15\% of the dataset pairs for validation and 5\% for testing. We selected this dataset for its high variety and diversity of concepts, as well as its prior use in T2I-Adapter training, making it particularly suitable for evaluating the controllability and fine-grained manipulation capabilities of our method.

\subsection{Evaluation Metrics}
To evaluate performance, we employ a set of metrics focused on controllability, image quality, and fidelity to user-defined regions. To ensure that overall image attributes are preserved during training, we use the Fréchet Inception Distance (FID) \cite{FID} and CLIPScore \cite{clipscore}, which measure image quality and image-text alignment, respectively. 

To specifically assess the accuracy of local control, we have modified the traditional Mean Squared Error (MSE) \cite{mse} to measure the precision of structural details within user-defined regions, referring to this adaptation as Downsampled Mean Squared Error (DMSE). In edge maps, generative models may produce contours that closely resemble the original structure but are slightly shifted in position. Consequently, direct pixel-wise comparisons can be misleading, penalizing even perceptually similar edges. To address this, we downsample the spatial condition (edges) and apply a Gaussian filter to enlarge the contours, increasing the likelihood of overlap between similar but slightly misaligned structures. This provides a more robust and tolerant measure of edge similarity. DMSE also enables a meaningful comparison between regions inside and outside the mask: a lower error inside the mask means the model adheres more closely to the spatial condition, while outside the mask, we compare the predicted latent $\widehat{z}_0$ to a latent vector of ones to analyze the amount of contour information generated in unconstrained areas.

Within this framework, two measures are reported: 1) $\text{DMSE}\text{in}$, the DMSE value inside the mask, where lower values mean higher structural similarity to the target region; and 2) $\text{DMSE}\text{out}$, the DMSE value outside the mask, where lower values indicate that more contours (edges) are being generated relative to a full ones latent vector, while higher values correspond to fewer generated edges. Importantly, $\text{DMSE}_\text{out}$ serves as an indicative value reflecting the amount of contour generation in unconstrained regions, rather than as a direct quality metric, as extreme values in either direction may indicate unrealistic outputs. This makes DMSE particularly useful for assessing not only the model’s fidelity within the masked region but also its generative behaviour in unconstrained areas.

\begin{figure*}
    \centering
    \begin{tabular}{c | 0{wc{0.15\textwidth}} | 0{wc{0.15\textwidth}} | 0{wc{0.15\textwidth}} | 0{wc{0.15\textwidth}} | 0{wc{0.15\textwidth}}}
         \toprule \toprule
		\noalign{\smallskip}
             \textbf{Sample} & \multicolumn{2}{c|}{\textbf{Original Image}} &
             \cellcolor[rgb]{0.918,0.663,0.290}\textbf{T2I-Adapter} & \cellcolor[rgb]{0.451,0.518,0.812}\textbf{ControlNet} & \cellcolor[rgb]{0.451,0.565,0.526}\textbf{Ours}\\
            \midrule
            \textit{a)} & \multicolumn{2}{c|}{\includegraphics[width=2.5cm, valign=c]{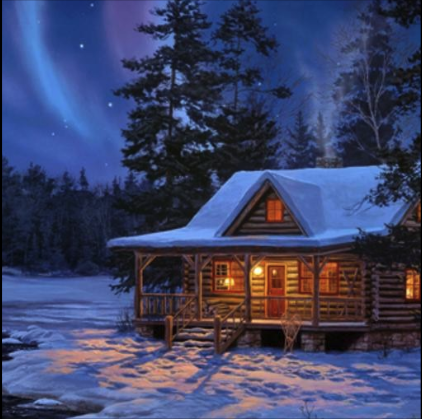}} &
            \includegraphics[width=2.5cm, valign=c]{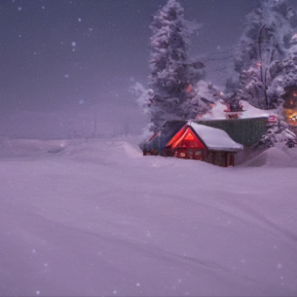} &
            \includegraphics[width=2.5cm, valign=c]{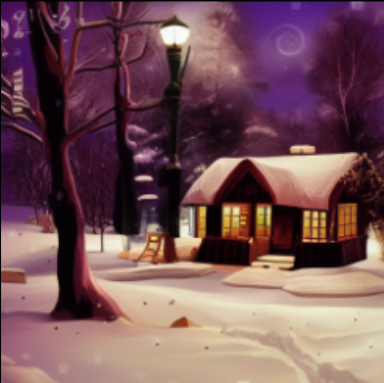} &
            \includegraphics[width=2.5cm, valign=c]{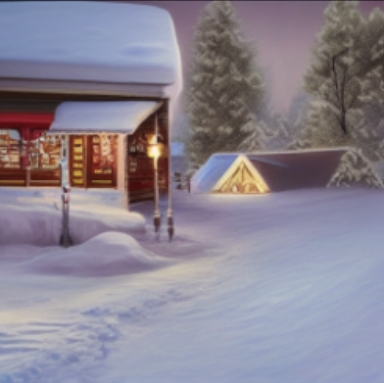} \\
            \cmidrule{2-6}
            \textit{b)} & \multicolumn{2}{c|}{\includegraphics[width=2.5cm, valign=c]{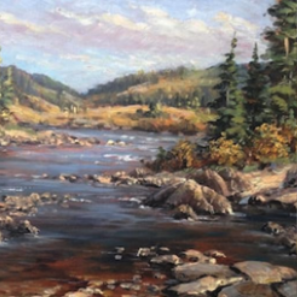}} &
            \includegraphics[width=2.5cm, valign=c]{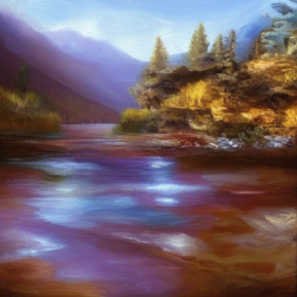} &
            \includegraphics[width=2.5cm, valign=c]{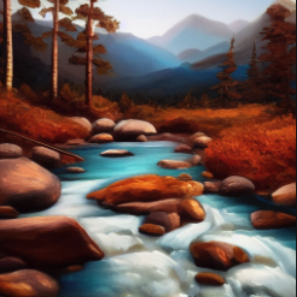} &
            \includegraphics[width=2.5cm, valign=c]{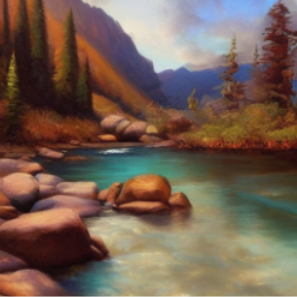} \\
            \midrule
            & \textbf{Mask} & \textbf{Regions} & \multicolumn{3}{c}{ } \\
            \midrule
            \textit{a)} & \multirow{ 2}{*}{\includegraphics[width=2.75cm, valign=c]{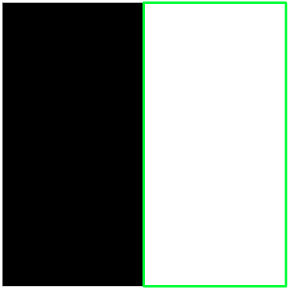}}&
            \includegraphics[width=2.5cm, valign=c]{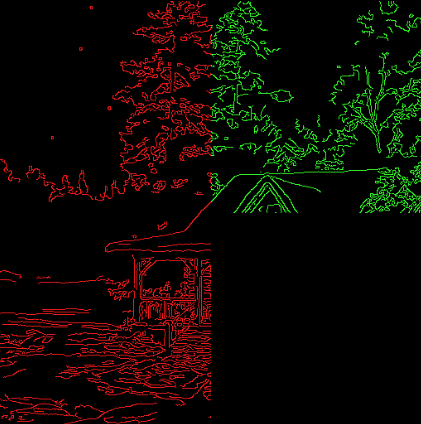}&
            \includegraphics[width=2.5cm, valign=c]{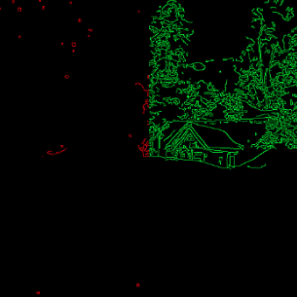}&
            \includegraphics[width=2.5cm, valign=c]{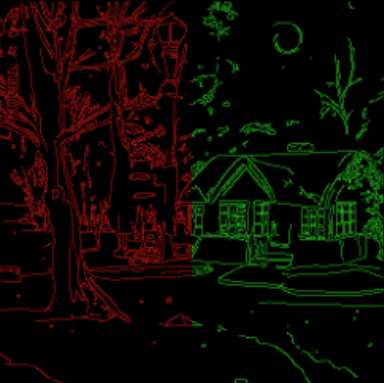}&
            \includegraphics[width=2.5cm, valign=c]{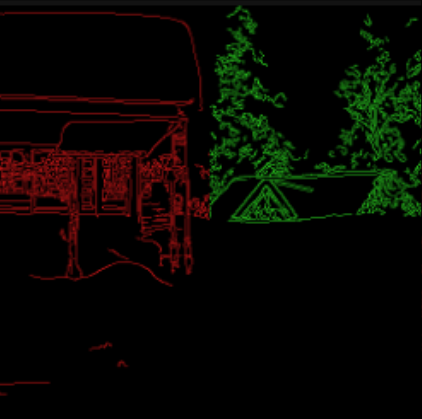}\\
            \cmidrule{3-6}
            \textit{b)} & &
            \includegraphics[width=2.5cm, valign=c]{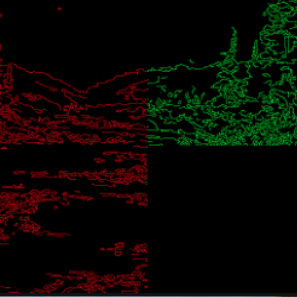}&
            \includegraphics[width=2.5cm, valign=c]{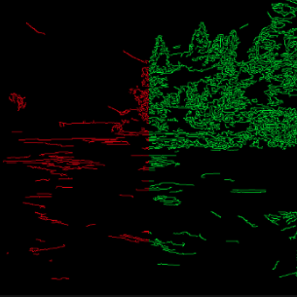}&
            \includegraphics[width=2.5cm, valign=c]{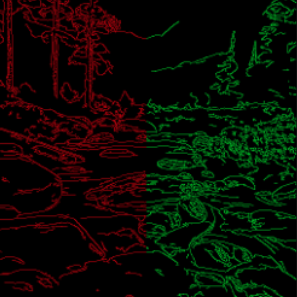}&
            \includegraphics[width=2.5cm, valign=c]{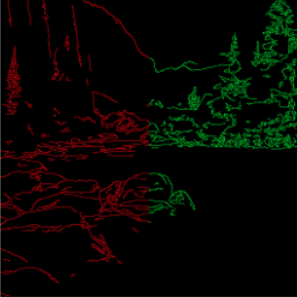}\\
            \midrule
            \multirow{ 2}{*}{\textit{a)}} & \multicolumn{2}{c|}{\textbf{$\text{DMSE}_\text{in} (\downarrow)$}} & 0.0062 & 0.0137 & 0.0048 \\
            \cmidrule{2-6}
            & \multicolumn{2}{c|}{\textbf{$\text{DMSE}_\text{out}$}} & 0.9925 & 0.8093 & 0.8870 \\
            \midrule
            \multirow{ 2}{*}{\textit{b)}} & \multicolumn{2}{c|}{\textbf{$\text{DMSE}_\text{in} (\downarrow)$}} & 0.0102 & 0.0163 & 0.0039 \\
            \cmidrule{2-6}
            & \multicolumn{2}{c|}{\textbf{$\text{DMSE}_\text{out}$}} & 0.9534 & 0.8311 & 0.9013 \\
            \midrule \bottomrule
    \end{tabular}
    \caption{Comparison between T2I-Adapter, ControlNet, and our proposed model for two different  images with prompts a) “\textit{A tavern in the cold winter night}” and b) "\textit{A landscape with a river, stones and some trees in oil painting}". The figure shows the original images from which edges were extracted, as well as the mask used to determine the ROI (the right half of the edges), which is used as input for all models. The “Regions” column displays the regions of the original edge map, where the bottom-right portion is set to zero, leaving only the upper-right with the original edges. The subsequent columns present the generated images and their corresponding edge maps for each model. The values of $\text{DMSE}\text{in}$ and $\text{DMSE}\text{out}$ are also shown for these examples, where (\textcolor{ForestGreen}{green areas}) indicate detected edges inside the ROI, and (\textcolor{red}{red areas}) indicate detected edges outside of it}
    \label{fig:visual_comparison}
\end{figure*}

Mathematically, we define $\text{DMSE}{\text{in}}$ and $\text{DMSE}{\text{out}}$ as shown in equations \eqref{eq:dmse_in} and \eqref{eq:dmse_out}. Where, $D(\cdot)$ represents a downsampling by a factor of 8, using a Gaussian filter with a $11 \times 11$ kernel and a sigma of 5, that is used to downsample both ground truth $E^{gt}$ and generated  $E^{gen}$ edge maps.
\begin{align}
    \text{DMSE}_{\text{in}} &= \frac{1}{\lVert M \rVert_{1}} 
    \sum_{i,j} (M_{i,j}) \cdot \notag  \\
    & \quad \left(D(E^{gt})_{i,j} - D(E^{gen})_{i,j}\right)^2 \label{eq:dmse_in} \\[0.5em]
    \text{DMSE}_{\text{out}} &= \frac{1}{\lVert 1 - M \rVert_{1}} 
    \sum_{i,j}(1 - M_{i,j}) \cdot \notag\\
    & \quad \left(1_{i,j}- D(E^{gen})_{i,j}\right)^2 \label{eq:dmse_out}
\end{align}
\subsection{Implementation Details}
For the implementation, Python \cite{python} version 3.12 was used alongside PyTorch \cite{pytorch} version 2 and the up-rising Diffusers library \cite{diffusers_library} to incorporate the proper modifications for our methods. Training was conducted on a single NVIDIA A10G Tensor Core GPU with 24GB of memory.

\begin{table*}
    \begin{center}
        \caption{Quantitative evaluation of T2I-Adapter, ControlNet, and our proposed method on 5,000 samples under the conditions of the \textit{Quadrant-Control Experiment}. Reported metrics include FID, CLIPScore, and $\text{DMSE}\text{in}$. Lower FID and $\text{DMSE}\text{in}$ values indicate better image quality and greater fidelity to the controlled area, respectively, while higher CLIPScore values reflect better image-text alignment.}
        \begin{tabular}{c | c | c | c | c }
            \toprule \toprule
		\noalign{\smallskip}
            \multirow{2}{*}{\textbf{Model}} & \multicolumn{3}{c|}{\textbf{With Prompt}} & \textbf{Without Prompt}\\
            & \textbf{FID ($\downarrow$)} & \textbf{CLIPScore ($\uparrow$)} & \textbf{$\text{DMSE}_\text{in}$ ($\downarrow$)} & \textbf{$\text{DMSE}_\text{in}$ ($\downarrow$)}  \\
            \noalign{\smallskip}
            \midrule
            \noalign{\smallskip}
            \textbf{T2I-Adapter} & 28.64 & 31.31 & 0.0086 $\pm$ 0.0081 & 0.0060 $\pm$ 0.0090 \\
            \textbf{ControlNet} & 26.63 & \textbf{32.67} & 0.0148 $\pm$ 0.0125 &
            0.0112 $\pm$ 0.0098\\
            \textbf{Ours} & \textbf{25.93} & 31.84 & \textbf{0.0066 $\pm$ 0.0069} & \textbf{0.0038 $\pm$ 0.0045} \\
            \midrule \bottomrule
        \end{tabular}
        \label{tab:comparison}
    \end{center}
\end{table*}

We employed the Adam optimizer with weight decay \cite{adamw} set to $1{\times}10^{-2}$, a learning rate of $8{\times}10^{-5}$ and a $\lambda$ value of $1{\times}10^{-3}$, determined through an ablation study described on \ref{sec:hyperparameters}. Models were trained with a batch size of 4 and gradient accumulation of 2, effectively simulating a batch size of 8. Training spanned up to 10 epochs with early stopping based on FID scores, evaluated every 10k steps. However, all models converged within a single epoch as FID scores ceased improving. We attribute this rapid convergence to the compact size of the T2I-Adapter and the use of a pre-trained U-Net.

\subsection{Results}\label{sec:results}
In this section, we evaluate our proposed method against two benchmarks: 1) a baseline T2I-Adapter trained with full Canny edges as conditioning; and 2) ControlNet trained under the same conditions. Our comparison focuses on two main criteria: 1) how well each model adheres to the designated user-defined area; and 2) how much edge structure is generated outside the mask in unconstrained areas.

To this end, we designed a controlled experiment that we call \textit{Quadrant-Control Experiment} in which, for each test image, the right half is designated as the ROI, while the left half is blacked out and serves as an unconditioned area. Within the ROI, the upper-right quadrant retains its original contours, while the lower-right quadrant is set to zero, to see how models act in those areas as no edges should be generated there. This setup allows us to simultaneously analyze $\text{DMSE}{\text{in}}$ (to measure the model’s generative behavior in unconstrained regions) and $\text{DMSE}{\text{out}}$ to measure the model’s generative behavior in unconstrained regions. For all methods, the input edge map is masked to define the ROI. However, in our method, the mask is also applied to the feature maps of the adapter.

Figure \ref{fig:visual_comparison} presents both a visual comparison and quantitative results for each method under the described experimental setup. In the figure, \textcolor{ForestGreen}{green areas} indicate contours within the target region, while \textcolor{red}{red areas} denote contours outside of it. This visualization allows for a direct comparison of how effectively each model maintains edge fidelity within the ROI and controls edge generation in unconditioned areas.

The results reveal two distinct trends. The baseline T2I-Adapter performs well within the ROI, preserving the desired contours, but struggles to generate contours outside the ROI, as indicated by high $\text{DMSE}\text{out}$ values, meaning an absence of contours. In contrast, ControlNet tends to generate a large number of edges across the entire image, including regions where no contours should appear, which results in higher $\text{DMSE}\text{in}$ scores due to poor adherence to the spatial condition. Notably, ControlNet often produces excessive edges even when the input condition is smooth. Our proposed method achieves high fidelity within the ROI, accurately following the edge condition, while also generating a moderate amount of contours outside the ROI, more than T2I-Adapter, but in a more controlled manner than ControlNet.

Since the structure and density of edges vary between samples, the absolute values of $\text{DMSE}\text{in}$ and $\text{DMSE}\text{out}$ also fluctuate from image to image. To capture the overall behavior of each method, we evaluated these metrics on 5,000 samples from the test partition, both with and without prompts, under the \textit{Quadrant-Control Experiment}. The results, summarized in Table \ref{tab:comparison}, report FID, CLIPScore, and $\text{DMSE}_\text{in}$. These results confirm that the observed trends are consistent across a diverse set of images, with our method demonstrating higher fidelity within the ROI.

Regarding performance outside the controlled region, and as discussed in the evaluation metrics section, $\text{DMSE}_\text{out}$ serves as an indicative measure of the number of contours generated, rather than a direct indicator of output quality or model preference. As shown in Table \ref{tab:dmse_out}, ControlNet generates the highest number of edges outside the ROI. However, a greater number of edges does not necessarily translate to better quality, as both our method and ControlNet achieve similar FID and CLIPScore values. Importantly, our method strikes a favorable balance, while it does not generate as many edges as ControlNet, it consistently achieves higher fidelity within the ROI and generates a more appropriate amount of edges outside of it, outperforming T2I-Adapter and demonstrating improved controllability and quality overall.

\begin{table}
    \caption{Comparison of $\text{DMSE}\text{out}$ values for T2I-Adapter, ControlNet, and our proposed method, evaluated on 5,000 test samples. $\text{DMSE}\text{out}$ serves as an indicative measure of the amount of contours generated outside the controlled area; higher or lower values do not directly reflect output quality, but help characterize each model’s generative behaviour in unconstrained areas.}
    \begin{tabular}{c | c | c }
        \toprule \toprule
    \noalign{\smallskip}
        \multirow{2}{*}{\textbf{Model}} & \textbf{With Prompt} & \textbf{Without Prompt}\\
        & \textbf{$\text{DMSE}_\text{out}$} & \textbf{$\text{DMSE}_\text{out}$}  \\
        \noalign{\smallskip}
        \midrule
        \noalign{\smallskip}
        \textbf{T2I-Adapter} & 0.9325 $\pm$ 0.0611 & 0.9601 $\pm$ 0.0666 \\
        \textbf{ControlNet} & 0.8646 $\pm$ 0.0960 & 0.8945 $\pm$ 0.0825\\
        \textbf{Ours} & 0.8805 $\pm$ 0.0814 & 0.9158 $\pm$ 0.0730 \\
        \midrule \bottomrule
    \end{tabular}
    \label{tab:dmse_out}
\end{table}

\subsection{Ablation Studies}
\label{sec:hyperparameters}
\begin{figure}
    \centering
    \begin{tabular}{{0{wc{0.15\textwidth}} | 0{wc{0.15\textwidth}} | 0{wc{0.15\textwidth}}}}
        \toprule \toprule
        \noalign{\smallskip}
        \textbf{Edges w/ROI} & \textbf{Gen. Edges} & \textbf{Gen. Image} \\
        \midrule
        \includegraphics[width=2.5cm, valign=c]{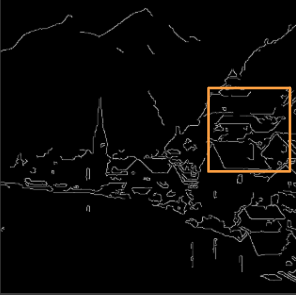} &
        \includegraphics[width=2.5cm, valign=c]{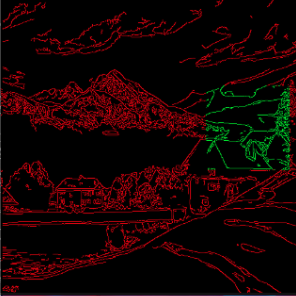} &
        \includegraphics[width=2.5cm, valign=c]{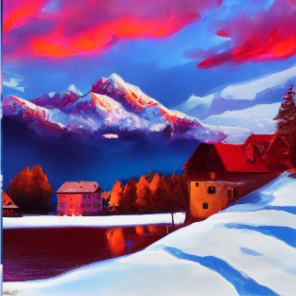} \\
        \midrule
        \includegraphics[width=2.5cm, valign=c]{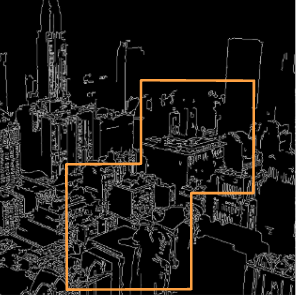} &
        \includegraphics[width=2.5cm, valign=c]{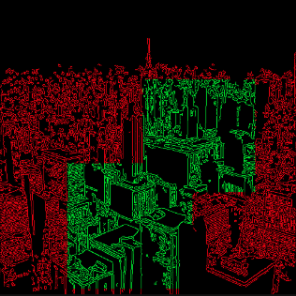} &
        \includegraphics[width=2.5cm, valign=c]{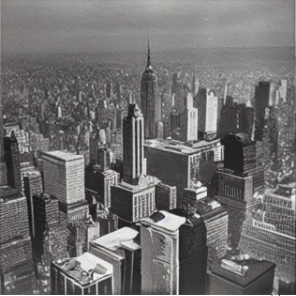} \\
        \midrule
        \includegraphics[width=2.5cm, valign=c]{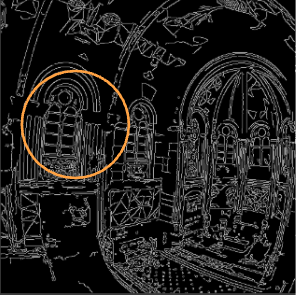} &
        \includegraphics[width=2.5cm, valign=c]{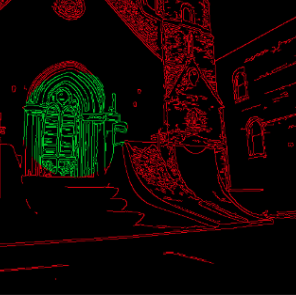} &
        \includegraphics[width=2.5cm, valign=c]{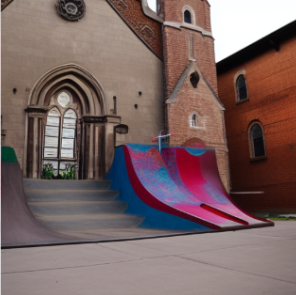} \\
        \midrule
        \includegraphics[width=2.5cm, valign=c]{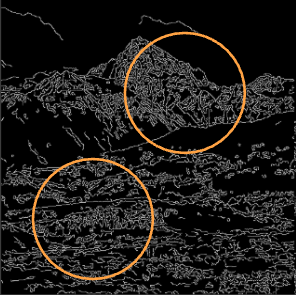} &
        \includegraphics[width=2.5cm, valign=c]{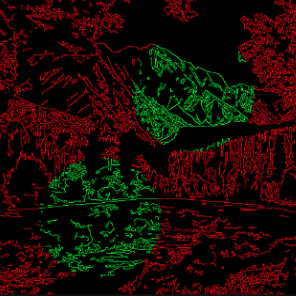} &
        \includegraphics[width=2.5cm, valign=c]{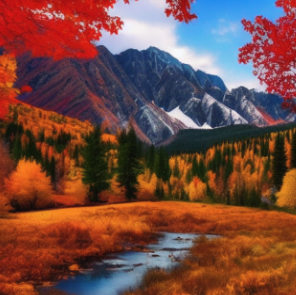} \\
        \midrule \bottomrule
    \end{tabular}
    \caption{Visualization of the localization ablation experiment. The first column displays the original edge map with the ROI highlighted in \textcolor{YellowOrange}{orange}. The second column shows the generated edges, where (\textcolor{ForestGreen}{green areas}) correspond to edges within the ROI and (\textcolor{red}{red areas}) indicate edges generated outside the ROI. The third column presents the final image synthesized by the model.}
    \label{fig:localization_ablation}
\end{figure}

To assess the robustness of our localization mechanism, we conducted experiments in which both the shape and the position of the mask were varied. This setup allows us to evaluate how well the model maintains spatial control under diverse masking conditions. Figure \ref{fig:localization_ablation} presents qualitative results demonstrating the model’s ability to adapt to different masks configurations.

Beyond localization, we also examined the impact of different filters used to extract edge information for the similarity loss. In particular, we compared the performance of the differentiable filters such as Sobel filter \cite{sobel} or Laplace. By testing both options, we aimed to determine which filter most effectively supports the learning of structural alignment in the conditioned area. The results of this comparison are summarized in Table \ref{tab:filter_ablation}, which reports performance with and without prompts under the \textit{Quadrant-Control Experiment} described in Section \ref{sec:results}. While the Sobel filter demonstrates a slight advantage over Laplace in terms of both image quality and edge fidelity within the ROI, the difference is minor. This suggests that either Sobel or Laplace can be effectively used to enforce structural similarity for our purposes.

\begin{table*}
    \begin{center}
        \caption{Quantitative evaluation of our method under different high pass filters (Sobel and Laplace) applied to 5,000 samples from the \textit{Quadrant-Control Experiment}. Reported metrics include FID, CLIPScore, and $\text{DMSE}_\text{in}$.}
        \begin{tabular}{c | c | c | c | c }
            \toprule \toprule
		\noalign{\smallskip}
            \multirow{2}{*}{\textbf{Model}} & \multicolumn{3}{c|}{\textbf{With Prompt}} & \textbf{Without Prompt}\\
            & \textbf{FID ($\downarrow$)} & \textbf{CLIPScore ($\uparrow$)} & \textbf{$\text{DMSE}_\text{in}$ ($\downarrow$)} & \textbf{$\text{DMSE}_\text{in}$ ($\downarrow$)}  \\
            \noalign{\smallskip}
            \midrule
            \noalign{\smallskip}
            \textbf{Sobel} & \textbf{25.93} & \textbf{31.84} & \textbf{0.0066 $\pm$ 0.0069} & \textbf{0.0038 $\pm$ 0.0045} \\
            \textbf{Laplace} & 26.51 & 31.68 & \textbf{0.0066 $\pm$ 0.0069} &
            0.01042 $\pm$ 0.0046\\
            \midrule \bottomrule
        \end{tabular}
        \label{tab:filter_ablation}
    \end{center}
\end{table*}

The $\lambda$ parameter plays a critical role in balancing penalization to enhance fidelity within the masked region. A $\lambda$ value of 0 results in training only the adapter by masking its features, while higher values introduce edge penalization within the mask into the loss function. To investigate the effects of $\lambda$ and the learning rate ($lr$), we conducted ablation experiments using the controlled setup in which half of each image is masked out. For each tested configuration, we generated 1,000 images from the test partition, with models trained for only 10,000 steps to observe general trends rather than to achieve full convergence. This approach allowed us to examine how different $\lambda$ and $lr$ values impact both fidelity inside the ROI and generative creativity outside. We found that while the metrics tend to improve gradually with additional training, most observable trends are established early, and the overall effects remain relatively consistent over time.

Our results, summarized in Table \ref{tab:ablation}, indicate that while there is no clear pattern suggesting that lowering either $\lambda$ or the learning rate ($lr$) consistently increases fidelity within the ROI, some trends can be observed. Specifically, lower learning rates tend to yield slightly higher fidelity, but the improvement is modest. Among all tested configurations, only three combinations outperformed the baseline with $\lambda = 0$. Notably, the best performance was achieved with a learning rate of $8{\times}10^{-5}$ and a $\lambda$ value of $1{\times}10^{-3}$. Although these improvements are incremental, they tend to become more apparent with additional training, highlighting that for instance, the best model trained for 50k iterations achieves a $\text{DMSE}_\text{in}$ of 0.066, compared to 0.086 for the $\lambda = 0$ training for the same iterations, demonstrating the effectiveness of including a small penalization term for local structure alignment.

\begin{table}
    \caption{Generative behaviour inside the controlled area. Results of varying ($\lambda$) and learning rate ($lr$) after 10k iterations, evaluated under the \textit{Quadrant-Control Experiment}. Only the mean values are considered as  they provide orientation for which method produces higher fidelity on average.}
    \begin{tabular}{c | c c c c c |}
    \toprule \toprule
    \noalign{\smallskip}
    \multicolumn{6}{c|}{$\text{DMSE}_\text{in}$ ($\downarrow$) on 1,000 samples} \\
    \midrule
    \multirow{ 2}{*}{$\lambda$} & \multicolumn{5}{c|}{$lr$} \\
    \multicolumn{1}{c|}{} & 1e-5 & 1e-4 & 4e-4 & 8e-4 & 8e-5 \\
    \noalign{\smallskip}
    \midrule
    \noalign{\smallskip}
    0 & 0.0101 & - & - & - & - \\
    10e-4 & 0.0119 & 0.0117 & - & - & \textbf{0.0092} \\
    25e-4 & - & - & 0.0114 & 0.0123 & 0.0132 \\
    50e-4 & - & - & 0.0097 & 0.0103 & 0.0095 \\
    75e-4 & 0.0119 & - & 0.0253 & - & - \\
    \midrule \bottomrule
    \end{tabular}
    \label{tab:ablation}
\end{table}

\section{Conclusions}\label{conclusions}
In this work, we introduced key innovations, such as including the use of masks and the use of a novel loss function into the latent space, to achieve localized control in diffusion models. These advancements significantly enhance the ability to manipulate specific regions of an image while maintaining overall image quality and ensuring the generation of coherent, complete images from partial information. Our contributions advance the field of text-to-image synthesis by enabling fine-grained local control, which has been challenging to achieve with previous methods.

Looking ahead, there is substantial potential to generalize our approach by introducing new conditions, such as color, and further exploring the use of attention maps for localized conditioning. Recent research has shown promising results with attention mechanisms \cite{ipadapter, gligen}, and incorporating conditions directly into these maps could offer a more robust and flexible method for controlling specific areas of an image. Notably, our experiments indicate that improving localized control does not compromise image-text alignment, as CLIPScore remains consistent across methods. Future work could investigate integrating semantic or hierarchical constraints—combining local spatial control with more sophisticated textual or conceptual guidance—to further expand the expressive power and precision of diffusion models.

%\bibliography{sn-bibliography}
%\bibliography
\printbibliography

\end{document}